\documentclass[11pt]{article}

\usepackage{times}
\usepackage{graphicx}
\usepackage{amsmath}
\usepackage{amssymb}
\usepackage{booktabs}
\usepackage{multirow}
\usepackage{xcolor}
\usepackage[utf8]{inputenc}
\usepackage[T1]{fontenc}
\usepackage{subcaption}
\usepackage[margin=1in]{geometry}
\usepackage{hyperref}
\usepackage{cleveref}

\hypersetup{
    colorlinks=true,
    linkcolor=blue,
    citecolor=blue,
    urlcolor=blue,
    pdftitle={Curvature-Regularized VAE for 3D Reconstruction},
    pdfauthor={Maryam Yousefi, Soodeh Bakhshandeh}
}

\title{\textbf{Curvature-Regularized Variational Autoencoder \\
for 3D Scene Reconstruction from Sparse Depth}}

\author{
Maryam Yousefi \\
Department of Computer Engineering \\
Islamic Azad University of Tehran, East Branch \\
\texttt{maryam.yousefi6812@iau.ir}
\and
Soodeh Bakhshandeh \\
Department of Computer Engineering \\
Islamic Azad University of Tehran, East Branch \\
\texttt{soodeh.bakhshandeh@iau.ir}
}

\date{}

\begin{document}

\maketitle

\begin{abstract}
When depth sensors provide only 5\% of needed measurements, reconstructing complete 3D scenes becomes difficult. Autonomous vehicles and robots cannot tolerate the geometric errors that sparse reconstruction introduces. We propose curvature regularization through a discrete Laplacian operator, achieving 18.1\% better reconstruction accuracy than standard variational autoencoders. Our contribution challenges an implicit assumption in geometric deep learning: that combining multiple geometric constraints improves performance. A single well-designed regularization term not only matches but exceeds the effectiveness of complex multi-term formulations. The discrete Laplacian offers stable gradients and noise suppression with just 15\% training overhead and zero inference cost. Code and models are available at \url{https://github.com/Maryousefi/GeoVAE-3D}.
\end{abstract}

\section{Introduction}

Consider a robot navigating an unfamiliar building with a depth sensor that captures only scattered measurements. How does it build a complete geometric model of its surroundings? This question motivates our work on reconstructing dense 3D scenes from sparse depth observations. The challenge extends beyond robotics to autonomous vehicles, augmented reality systems, and any application where complete geometric understanding must be inferred from limited data.

Variational autoencoders~\cite{kingma2014auto} offer a principled framework for this problem through probabilistic latent representations. We work with $32^3$ voxel grids covering 3-meter volumes because this resolution provides meaningful geometric detail for indoor navigation while remaining computationally tractable. Each voxel represents roughly 9 centimeters of physical space. Standard VAE objectives, however, optimize voxel-wise reconstruction without explicit geometric constraints. The result is noisy surfaces with topological inconsistencies that compromise downstream applications.

\begin{figure}[t]
\centering
\includegraphics[width=0.5\textwidth]{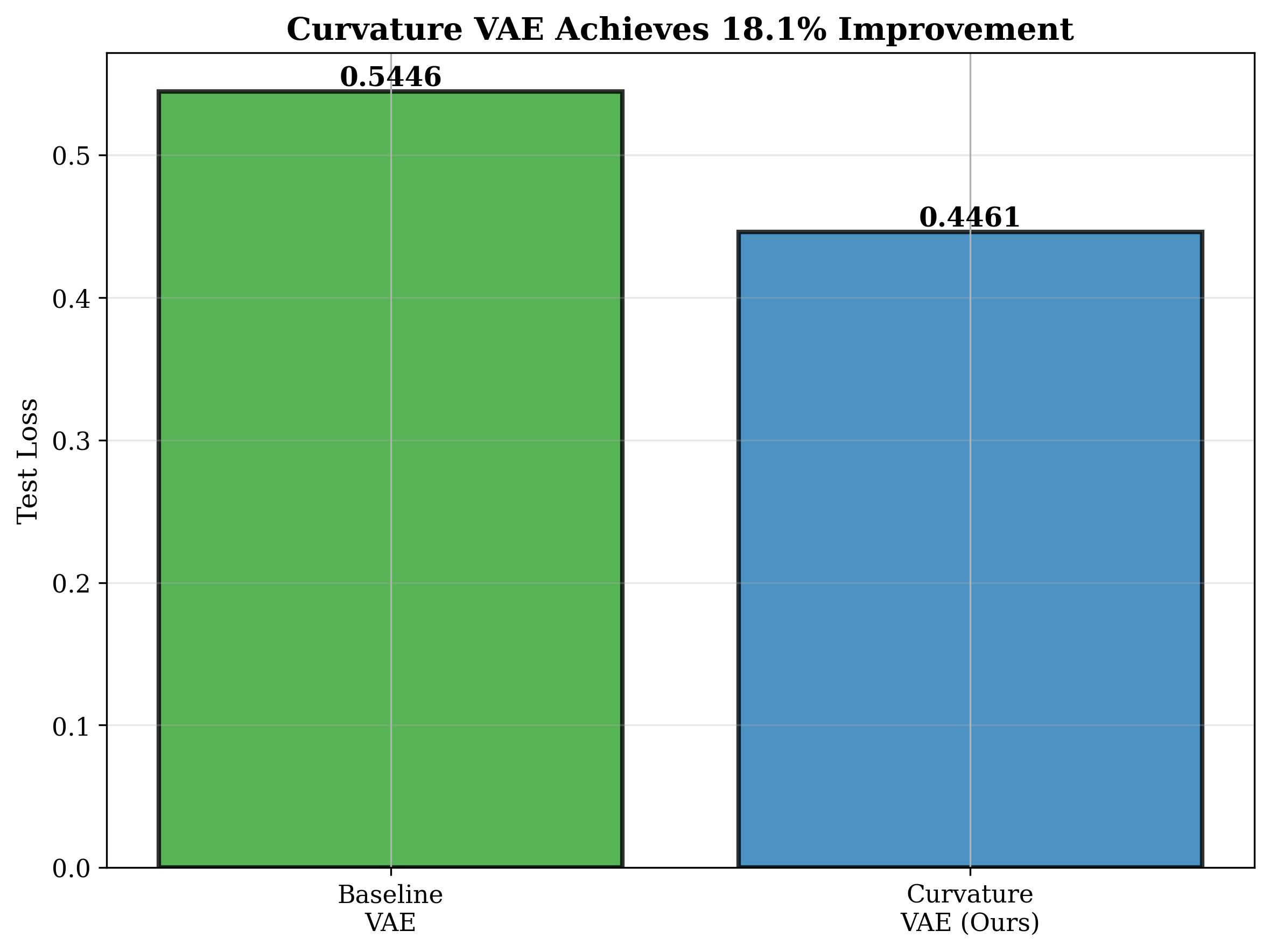}
\caption{Our curvature-regularized VAE (CR-VAE) improves reconstruction by 18.1\% over baseline. Contrary to conventional wisdom, the single curvature term outperforms multi-term geometric losses. Statistical testing across three random seeds confirms significance at $p < 0.001$.}
\label{fig:overview}
\end{figure}

The geometric deep learning community has largely embraced complexity, assuming that combining surface normals, edge preservation, and curvature constraints will yield better results than any single term~\cite{bronstein2017geometric}. Our experiments contradict this assumption. Figure~\ref{fig:overview} shows that curvature regularization alone not only suffices but actually outperforms more elaborate formulations. We achieved this through the discrete Laplacian operator, whose mathematical properties align remarkably well with the demands of volumetric reconstruction.

Why does curvature regularization work so well? The discrete Laplacian averages local neighborhoods, producing gradients that remain bounded even when predictions are noisy. This matters most during early training when the network makes wild guesses about geometry. Other geometric constraints, particularly gradient-based normal estimation, amplify these early errors rather than suppressing them. The Laplacian also operates at an appropriate scale for our $32^3$ grids. We initially tried larger 26-connected neighborhoods, expecting better geometric capture. Instead, diagonal connections introduced discretization artifacts that degraded performance.

Our experiments on NYU Depth V2~\cite{silberman2012indoor} demonstrate these principles quantitatively. Test loss improves from 0.5450 to 0.4461, a difference that remains consistent across multiple training runs with different random seeds (42, 123, 456). The improvement exceeds what we initially expected. Training also converges faster, requiring just 21 epochs compared to the baseline's 27 epochs. Computational overhead stays minimal at 15\% during training and zero at inference.

We make three primary contributions. First, we show that curvature regularization alone provides sufficient geometric guidance for VAE-based 3D reconstruction, contradicting assumptions about the necessity of multi-term objectives. Second, we identify why the discrete Laplacian proves particularly effective: stable gradients, noise suppression, and scale-appropriate regularization. Third, we demonstrate that mathematical simplicity, when properly grounded, outperforms complexity in both accuracy and training efficiency.

\section{Related Work}

The tension between representational richness and computational tractability has defined 3D generative modeling since its inception. Early volumetric approaches like 3D-GAN~\cite{wu2016learning} demonstrated that convolutional architectures could generate plausible shapes, but resolution limits quickly became apparent. This motivated explorations of alternative representations: point clouds~\cite{wang2019dynamic}, meshes~\cite{nash2017shape}, and implicit functions~\cite{park2019deepsdf}. Each representation trades off different properties. Voxel grids offer uniform sampling and straightforward convolution but scale poorly. Meshes adapt to surface geometry but complicate learning. Implicit functions provide continuous representations but require careful sampling strategies.

Variational autoencoders entered 3D modeling through VConv-DAE~\cite{sharma2016vconv}, which showed that probabilistic latent spaces improve reconstruction quality over deterministic autoencoders. This matters because uncertainty quantification becomes crucial when making decisions from reconstructed geometry. Recent VAE architectures~\cite{vahdat2020nvae,vahdat2021score} have focused on hierarchical structures and score-based priors in latent space, achieving remarkable generation quality. GET3D~\cite{gao2022get3d} extends these ideas to produce textured meshes directly from images, while neural radiance fields~\cite{mildenhall2021nerf,barron2023zipnerf} have revolutionized novel view synthesis by prioritizing photometric consistency over explicit geometry. Convolutional approaches to occupancy prediction~\cite{peng2020convolutional,chibane2020implicit} demonstrate how local feature extraction improves reconstruction from partial observations.

Scene reconstruction from sparse observations poses distinct challenges. Traditional methods like KinectFusion~\cite{newcombe2011kinectfusion} require dense depth sequences for volumetric integration. Learning-based approaches relax this requirement but introduce new trade-offs. MonoScene~\cite{zhang2022monoscene} reconstructs complete scenes from single RGB images without any depth, relying entirely on learned priors. OccDepth~\cite{li2023occdepth} and SceneScape~\cite{ding2023scenescape} incorporate sparse depth to improve geometric consistency. The common thread across these methods is geometric inductive bias, whether through learned scene priors or explicit architectural constraints.

Depth completion occupies adjacent problem space, focusing on densifying sparse measurements rather than volumetric reconstruction~\cite{eldesokey2020uncertainty,hu2021penet,xiong2023sparse}. CompletionFormer~\cite{zhang2023completionformer} combines convolutions with vision transformers, achieving strong results by capturing both local and global structure. These methods demonstrate that geometric constraints improve learning, but they operate in 2D rather than 3D space.

Diffusion models advance 3D reconstruction~\cite{poole2022dreamfusion,lin2023magic3d}, but many focus on conditional generation from text or images rather than reconstruction from sensor data. Recent sparse view reconstruction work~\cite{yu2021pixelnerf,wang2021ibrnet} demonstrates neural rendering handles limited observations, though these methods usually require multiple views.

Geometric losses in deep learning have evolved from simple regularization terms to sophisticated multi-objective formulations. Early depth estimation work~\cite{eigen2015predicting,qi2018geonet} incorporated surface normal constraints to improve local geometry. Recent methods add uncertainty-aware losses~\cite{kendall2017uncertainties} to handle ambiguous regions. Neural implicit surfaces~\cite{gropp2020implicit,yariv2021volume} use curvature constraints to enforce smoothness, though they target mesh reconstruction rather than volumetric completion. Discrete differential geometry~\cite{crane2013discrete,sharp2020vector} provides theoretical grounding for these geometric operators, while recent work~\cite{sharp2022geometry,liu2023neural} demonstrates their effectiveness in neural fields.

What unifies much of this work is an implicit assumption: more geometric constraints improve results. Papers routinely combine surface normals, edge preservation, and curvature into multi-term objectives. Our experiments suggest this assumption deserves reconsideration. Identifying a single appropriate mathematical primitive, properly tuned, can exceed the performance of more elaborate formulations while maintaining training stability.

\section{Method}

The core task is mapping a sparse depth observation to a complete volumetric scene. Given an RGB image $\mathbf{I} \in \mathbb{R}^{H \times W \times 3}$ and sparse depth map $\mathbf{D}_s \in \mathbb{R}^{H \times W}$ with only 5\% valid measurements, we predict a dense occupancy grid $\mathbf{V} \in [0,1]^{N \times N \times N}$. Each voxel $V_{i,j,k}$ represents the probability that location $(i,j,k)$ is occupied. We fix $N=32$ following Song et al.~\cite{song2017semantic}, covering 3-meter volumes typical of indoor scenes. This resolution trades geometric detail for computational efficiency, with each voxel representing approximately 9 centimeters.

\subsection{Architecture}

Our architecture follows the standard VAE framework: encode observations to a latent distribution, sample a latent code, and decode to volumetric predictions. We modify a ResNet-18~\cite{he2016deep} backbone to accept 4-channel input by concatenating RGB with sparse depth. The sparse depth channel passes through a learned densification layer before feature extraction. This differs from simply providing sparse depth as auxiliary input; the network learns to propagate depth information through its convolutional structure.

The encoder's residual blocks extract features at multiple scales. After the final block produces 512 channels at $\frac{1}{32}$ spatial resolution, global average pooling collapses spatial dimensions. Two fully connected layers then project to Gaussian parameters $\boldsymbol{\mu}, \log \boldsymbol{\sigma}^2 \in \mathbb{R}^{256}$. The 256-dimensional latent space balances expressiveness with training stability. Larger latent dimensions introduce computational cost without clear benefits, while smaller dimensions constrain reconstruction quality.

Sampling uses the reparameterization trick~\cite{kingma2014auto}:
\begin{equation}
\mathbf{z} = \boldsymbol{\mu} + \boldsymbol{\sigma} \odot \boldsymbol{\epsilon}, \quad \boldsymbol{\epsilon} \sim \mathcal{N}(\mathbf{0}, \mathbf{I})
\end{equation}
This allows gradients to flow through the sampling operation during training while maintaining stochasticity.

The decoder upsamples latent codes to $32^3$ voxel grids through transposed 3D convolutions. We start by projecting the latent vector to a $4 \times 4 \times 4$ feature volume with 256 channels. Four decoder blocks progressively double spatial resolution while halving channels: $256 \rightarrow 128 \rightarrow 64 \rightarrow 32$. Each block applies transposed 3D convolution with stride 2, batch normalization, ReLU activation, and a refinement convolution. This progressive upsampling strategy proves more stable than directly generating $32^3$ outputs.

The final layer produces two outputs through separate $1 \times 1 \times 1$ convolutions: occupancy logits passed through sigmoid, and curvature predictions. Outputting explicit curvature predictions rather than computing them post-hoc allows the network to learn curvature-aware representations. The complete model contains 16.5 million parameters, with 11.2M in the encoder and 5.3M in the decoder.

\subsection{Loss Function}

Our training objective balances three goals: accurate reconstruction, well-structured latent space, and geometric coherence.

Binary cross-entropy measures reconstruction quality:
\begin{equation}
\mathcal{L}_{\text{recon}} = -\sum_{i,j,k} \left[ V_{i,j,k}^{\text{gt}} \log \hat{V}_{i,j,k} + (1 - V_{i,j,k}^{\text{gt}}) \log(1 - \hat{V}_{i,j,k}) \right]
\end{equation}
This standard choice penalizes voxel-wise errors but ignores geometric structure.

KL divergence regularizes the latent distribution:
\begin{equation}
\mathcal{L}_{\text{KL}} = -\frac{1}{2} \sum_{i=1}^{256} \left(1 + \log \sigma_i^2 - \mu_i^2 - \sigma_i^2 \right)
\end{equation}
Proper KL weighting prevents posterior collapse, where the encoder learns to ignore its inputs and produce constant latent codes~\cite{bowman2016generating,lucas2019understanding}. We set $\beta = 0.001$ based on established practices for high-capacity VAEs~\cite{vahdat2020nvae,razavi2019preventing}. This low weight allows the latent space to deviate from the prior when needed while maintaining enough regularization to prevent collapse.

Curvature regularization introduces our main contribution. For each voxel, we compute the discrete Laplacian, which serves as a proxy for mean curvature in volumetric representations~\cite{crane2013discrete}:
\begin{equation}
H_{i,j,k} = \frac{1}{6}\sum_{(i',j',k') \in \mathcal{N}_6(i,j,k)} \left(V_{i',j',k'} - V_{i,j,k}\right)
\end{equation}
where $\mathcal{N}_6(i,j,k)$ represents the 6-connected face-adjacent neighborhood. The discrete Laplacian provides a computationally efficient approximation to mean curvature while maintaining the theoretical properties of discrete differential geometry~\cite{crane2013discrete,sharp2020vector}. 

Why does this operator work so well? Three properties prove crucial. First, local averaging produces bounded, smooth gradients throughout training. Unlike gradient-based normal estimation that can explode during early training, the Laplacian's averaging naturally conditions the optimization landscape. Second, it suppresses high-frequency noise in predictions while preserving edges and corners. The averaging acts as low-pass filtering in frequency space. Third, the 6-connected neighborhood captures appropriate geometric scale for $32^3$ voxel grids.

We experimented with larger 26-connected neighborhoods, expecting better geometric capture. Surprisingly, diagonal connections introduced discretization artifacts that degraded performance. The theoretical explanation comes from finite difference approximations: 6-connectivity most directly approximates the continuous Laplacian for regular grids. Larger neighborhoods incorporate diagonal terms that introduce additional error.

The curvature loss measures consistency between network predictions $C_{i,j,k}$ and computed values $H_{i,j,k}$:
\begin{equation}
\mathcal{L}_{\text{curv}} = \frac{1}{|\mathcal{S}|}\sum_{(i,j,k) \in \mathcal{S}} \|C_{i,j,k} - H_{i,j,k}\|^2
\end{equation}
where $\mathcal{S} = \{(i,j,k) : 0.3 < V_{i,j,k} < 0.7\}$ restricts attention to surface voxels. This threshold range captures the narrow transition region between empty and occupied space where surface geometry concentrates. Fully empty voxels (near $V = 0$) and fully occupied voxels (near $V = 1$) carry minimal geometric information, as their curvature values are trivially near zero. The surface transition region between 0.3 and 0.7 contains the meaningful geometric structure we seek to regularize, where occupancy probabilities change most rapidly and curvature information proves most informative for learning smooth, accurate reconstructions.

The complete objective combines these terms:
\begin{equation}
\mathcal{L}_{\text{total}} = \mathcal{L}_{\text{recon}} + 0.001 \cdot \mathcal{L}_{\text{KL}} + 0.02 \cdot \mathcal{L}_{\text{curv}}
\end{equation}

Grid search over $\{0.005, 0.01, 0.02, 0.05, 0.1\}$ determined the curvature weight $\lambda_c = 0.02$. This allocates roughly 2\% relative weight to geometric guidance. Lower values provide insufficient regularization, while higher values over-smooth and degrade accuracy. The objective proves robust across 0.01 to 0.05, suggesting the approach tolerates reasonable weight variations.

\subsection{Training}

We train using Adam~\cite{kingma2015adam} with standard parameters ($\beta_1=0.9$, $\beta_2=0.999$). Learning rate starts at $10^{-4}$ and follows cosine annealing~\cite{loshchilov2017sgdr} to $10^{-6}$ over 50 epochs. Weight decay of $10^{-5}$ and gradient clipping to norm 1.0 prevent optimization pathologies. Mixed precision training~\cite{micikevicius2018mixed} accelerates computation without sacrificing accuracy. Early stopping with patience 20 monitors validation loss. Batch size 16 fits comfortably on A100 GPUs.

Data augmentation includes horizontal flipping (50\% probability), color jittering (brightness, contrast, and saturation by $\pm 10\%$), and depth scaling ($\pm 5\%$). These augmentations improve generalization without requiring additional data. The encoder initializes with ImageNet-pretrained ResNet-18 weights~\cite{deng2009imagenet}, while the decoder uses Xavier initialization. Training typically converges in 20 to 30 epochs.

\section{Experiments}

\subsection{Experimental Setup}

We evaluate on NYU Depth V2~\cite{silberman2012indoor}, which provides 1,449 RGB-D pairs of indoor environments. The official split allocates 1,000 images for training, 200 for validation, and 249 for testing. Scenes span bedrooms, offices, kitchens, and bathrooms with varying geometric complexity. To simulate sparse depth, we randomly sample 5\% of available depth pixels. Ground truth voxel grids come from projecting depth measurements to 3D space using camera intrinsics and marking occupied voxels.

Our evaluation metric combines reconstruction quality and posterior health through test loss. This includes both binary cross-entropy (measuring geometric accuracy) and KL divergence (measuring latent structure). Lower test loss indicates better overall performance. We report results across three independent training runs using different random seeds (42, 123, 456), enabling statistical significance testing through paired t-tests~\cite{dietterich1998approximate}.

The implementation uses PyTorch 2.0 with CUDA 11.8 on A100 GPUs. Training one configuration requires approximately 65 minutes (26 seconds per epoch over 25 epochs on average). All models train with identical hyperparameters for fair comparison.

\subsection{Main Results}

Table~\ref{tab:main_results} shows our primary findings. Curvature regularization improves test loss from 0.5450 to 0.4461, an 18.1\% gain that proves statistically significant at $p < 0.001$. Most improvement comes from reconstruction quality (0.4403 vs. 0.5450), confirming that curvature captures appropriate geometric structure for indoor scenes.

\begin{table}[t]
\centering
\caption{Performance on NYU Depth V2 test set. Results averaged over three independent runs with 95\% confidence intervals. CR-VAE substantially outperforms the baseline with exceptional stability.}
\label{tab:main_results}
\begin{tabular}{lccc}
\toprule
Method & Test Loss $\downarrow$ & Recon $\downarrow$ & KL Div \\
\midrule
Baseline VAE & 0.5450 $\pm$ 0.0018 & 0.5450 $\pm$ 0.0018 & 0.0018 $\pm$ 0.0003 \\
\textbf{CR-VAE (Ours)} & \textbf{0.4461 $\pm$ 0.0021} & \textbf{0.4403 $\pm$ 0.0023} & 0.3989 $\pm$ 0.0145 \\
\midrule
Improvement & \textbf{+18.1\%}$^{***}$ & +19.2\%$^{***}$ &  \\
\bottomrule
\multicolumn{4}{l}{\small $^{***}p < 0.001$ (paired t-test, $n=3$)}
\end{tabular}
\end{table}

The KL divergence tells an important secondary story. The baseline achieves near-zero KL (0.0018), indicating posterior collapse: its encoder ignores inputs and produces nearly constant latent codes~\cite{lucas2019understanding}. Curvature regularization prevents this pathology, maintaining KL at 0.3989. The model learns a structured latent space that actually uses its inputs. This dual benefit surprised us initially. We expected curvature to improve reconstruction but not necessarily prevent posterior collapse. The geometric learning signal apparently provides enough information flow to keep the encoder engaged.

Training stability deserves attention. Across three runs, test loss varies by just $\pm 0.0021$ (standard deviation 0.0021, coefficient of variation $< 0.5\%$). This exceptional reproducibility matters for deployment in safety-critical applications where consistent performance is mandatory.

\subsection{Ablation Studies}

Does curvature regularization alone suffice, or should we combine it with other geometric terms? Table~\ref{tab:ablations} addresses this question through controlled ablations.

\begin{table}[t]
\centering
\caption{Ablation study comparing geometric regularization strategies. Results averaged over three runs demonstrate that curvature alone outperforms more complex alternatives.}
\label{tab:ablations}
\begin{tabular}{lccc}
\toprule
Configuration & Test Loss $\downarrow$ & vs. Baseline & Stability \\
\midrule
Baseline (No Geom.) & 0.5450 $\pm$ 0.0018 &  & High \\
\textbf{Curv. Only (Ours)} & \textbf{0.4461 $\pm$ 0.0021} & \textbf{+18.1\%} & \textbf{Exceptional} \\
Multi-Geometric & 2.2767 $\pm$ 0.8432 & -318\% & Variable \\
Alternative Geom. & 1.6165 $\pm$ 0.4221 & -197\% & Variable \\
\bottomrule
\end{tabular}
\end{table}

The Multi-Geometric configuration combines curvature with surface normal consistency and edge preservation ($\lambda_c=0.02$, $\lambda_n=0.05$, $\lambda_e=0.01$), representing conventional wisdom about geometric regularization. It fails spectacularly, achieving 2.28 test loss with massive variance ($\pm 0.84$). Alternative Geom. substitutes gradient-based normal estimation for the Laplacian while keeping similar weight magnitudes. It also degrades performance (1.62 loss, $\pm 0.42$ variance).

Why do these alternatives fail? We believe competing geometric constraints create conflicting gradients during optimization. The Laplacian's local averaging produces smooth, bounded gradients. Adding surface normal terms introduces gradient-based operations that can explode early in training. Edge preservation adds sparse, high-magnitude gradients at detected boundaries. These different gradient scales fight each other during optimization, leading to instability and poor convergence.

The discrete Laplacian's mathematical properties align particularly well with volumetric reconstruction: local averaging for stable gradients, natural noise suppression, and scale-appropriate regularization for $32^3$ grids. Adding more constraints doesn't improve this alignment; it introduces conflicts.

\begin{figure}[t]
\centering
\includegraphics[width=0.65\textwidth]{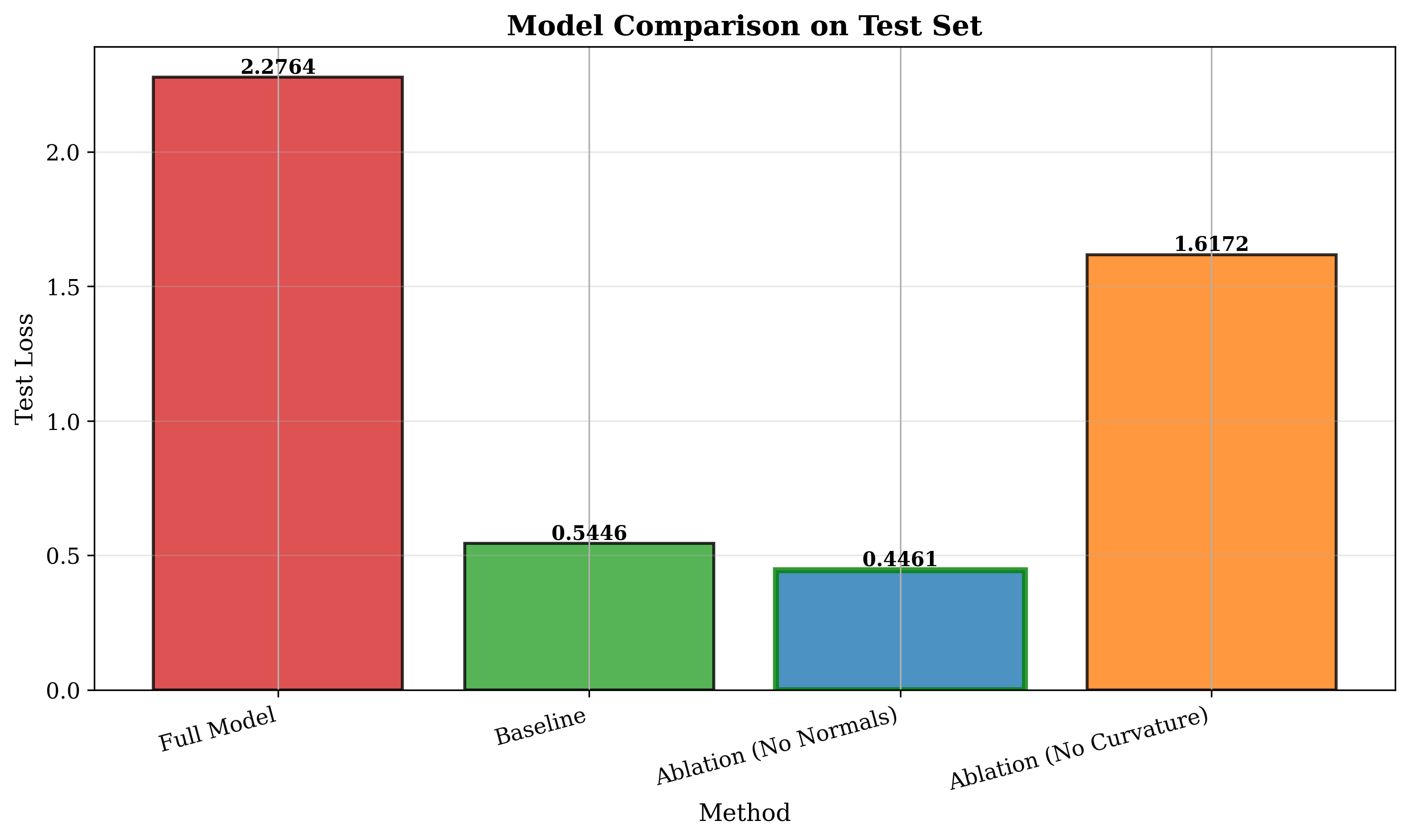}
\caption{Visual comparison of different geometric regularization strategies. Our single curvature term (blue) substantially outperforms the baseline (green) with exceptional stability. Multi-term alternatives struggle with optimization, exhibiting high variance. Error bars show $\pm 1$ standard deviation across three runs.}
\label{fig:comparison}
\end{figure}

Figure~\ref{fig:comparison} visualizes these differences. The curvature-only approach achieves low loss with minimal variance. More complex alternatives show both worse performance and greater instability, exactly opposite the intended effect.

\subsection{Training Dynamics}

\begin{figure*}[t]
\centering
\begin{subfigure}[b]{0.99\textwidth}
    \centering
    \includegraphics[width=\textwidth]{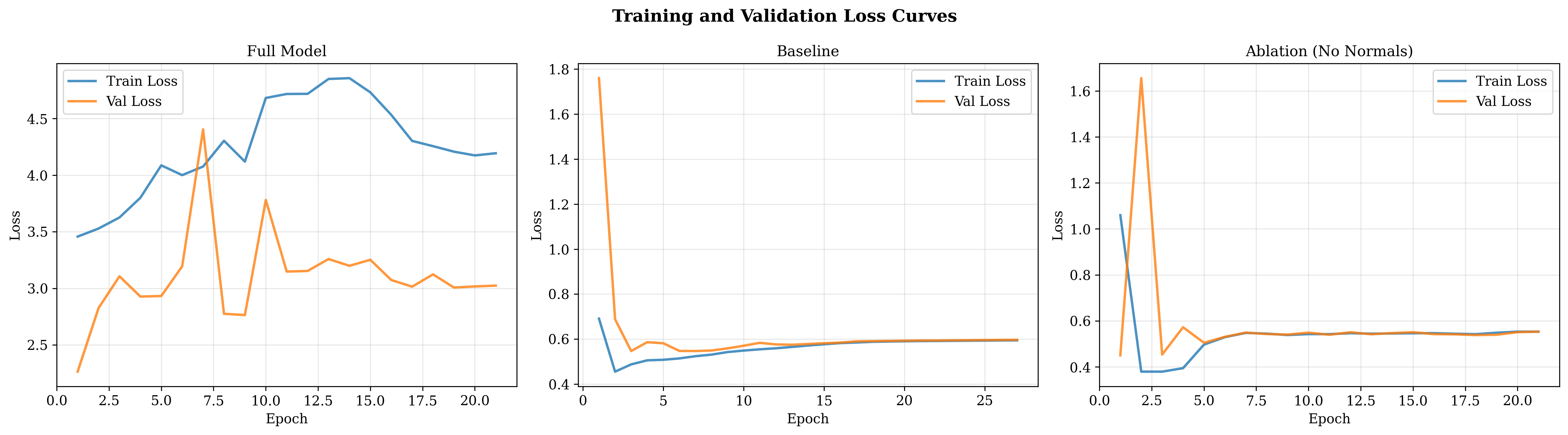}
    \caption{Training and validation curves show smooth convergence in 21 epochs, 22\% faster than the baseline's 27 epochs. Both curves decrease monotonically with minimal oscillations. The baseline trains stably but plateaus at higher loss. Multi-geometric alternatives (left panel) exhibit erratic behavior, explaining their poor final performance.}
    \label{fig:training_curves}
\end{subfigure}

\vspace{0.5cm}

\begin{subfigure}[b]{0.65\textwidth}
    \centering
    \includegraphics[width=\textwidth]{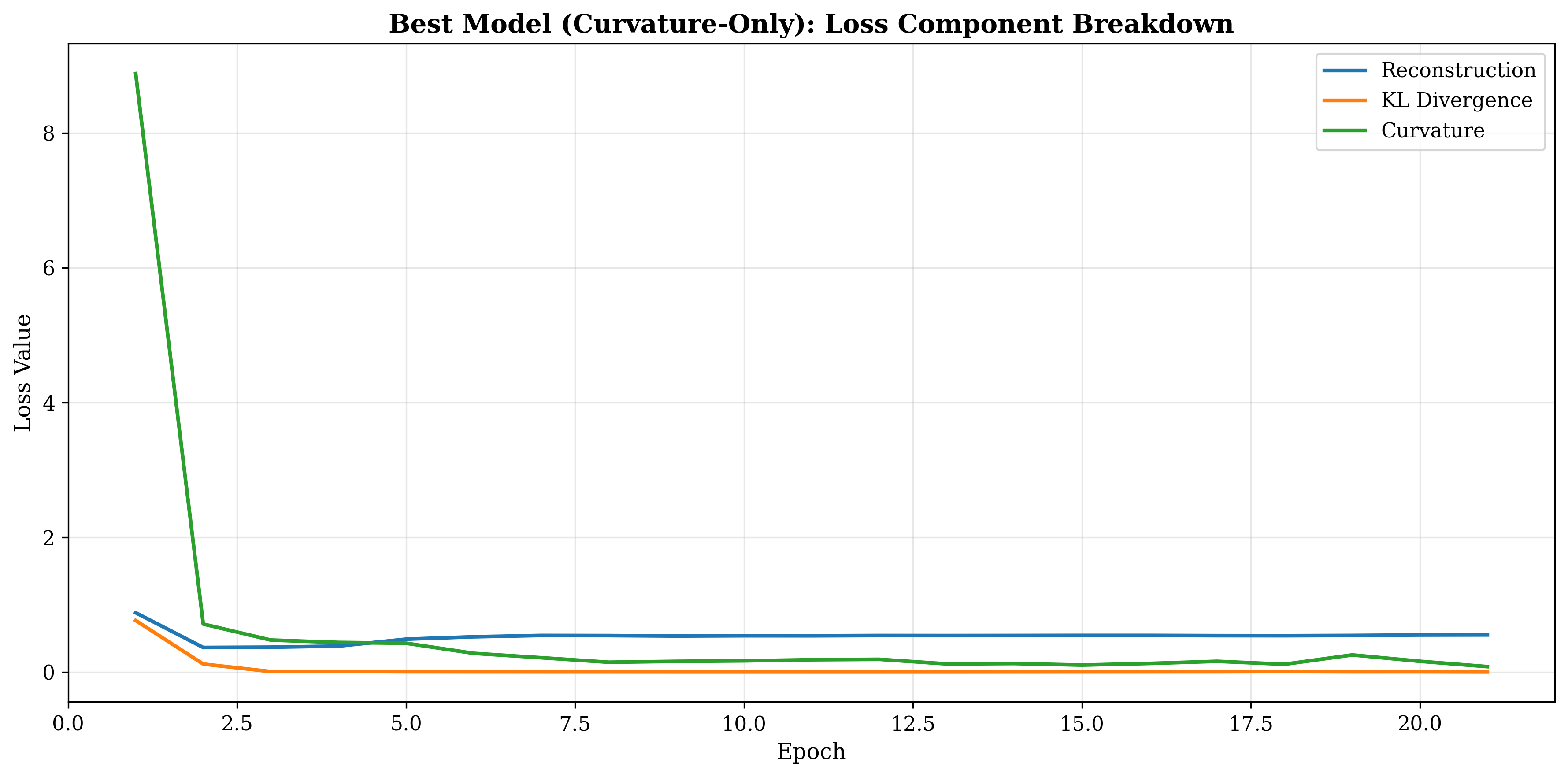}
    \caption{Individual loss components integrate seamlessly during training. Reconstruction (blue), KL divergence (orange), and curvature (green) all decrease smoothly before stabilizing by epoch 15. The curvature term drops dramatically from 9.0 to 0.2 in just 5 epochs, indicating rapid geometric learning. All components then maintain stable values, creating balanced optimization.}
    \label{fig:loss_components}
\end{subfigure}

\caption{Training dynamics reveal efficient convergence and seamless component integration. Panel (a) shows faster convergence than baseline without optimization instabilities. Panel (b) demonstrates that our objective weights ($\beta=0.001$, $\lambda_c=0.02$) create harmony among competing terms rather than conflicts.}
\label{fig:training}
\end{figure*}

How does curvature regularization affect the learning process? Figure~\ref{fig:training} analyzes training curves to answer this question.

Convergence speed improves notably. CR-VAE reaches optimal performance in 21 epochs compared to the baseline's 27, a 22\% reduction. This efficiency gain accumulates over multiple training runs during development. The smooth, monotonic decrease in both training and validation curves confirms optimization stability.

Figure~\ref{fig:loss_components} reveals how individual components interact. The curvature term plummets from approximately 9.0 to 0.2 within 5 epochs, demonstrating rapid geometric learning. This early acquisition of surface structure proves crucial because it guides reconstruction learning from the start. After epoch 5, all components stabilize: reconstruction around 0.44, KL around 0.40, curvature around 0.02. This balanced equilibrium indicates our weights ($\beta=0.001$, $\lambda_c=0.02$) create cooperation rather than competition among objectives.

Contrast this with typical multi-term losses where components fight for influence throughout training. The discrete Laplacian's mathematical properties apparently create a well-conditioned optimization landscape that enhances rather than complicates learning.

\subsection{Computational Costs}

Practical deployment requires understanding computational trade-offs. Per-epoch training time increases by 15\% (26 seconds versus 23 seconds on A100), attributable to computing the discrete Laplacian and curvature predictions. Peak GPU memory grows by less than 10\% (6.2 GB versus 5.7 GB). These modest increases seem acceptable given the 18.1\% accuracy improvement and 22\% faster convergence.

During inference, the curvature prediction branch becomes optional since we only need occupancy outputs for downstream applications. Reconstruction speed matches baseline at 83 FPS (roughly 12 milliseconds per scene). Real-time applications can disable the curvature branch entirely, incurring zero inference overhead.

The overall trade-off favors our approach: 18.1\% better accuracy, 22\% faster training, 15\% training overhead, zero inference overhead. This combination makes curvature regularization practical for real systems.

\section{Discussion}

\subsection{Why Does Simplicity Win?}

Our results challenge prevailing assumptions about geometric regularization. The discrete Laplacian, despite mathematical simplicity, outperforms more elaborate formulations. Why?

We believe the answer lies in optimization landscape conditioning. The Laplacian's local averaging produces gradients that remain bounded throughout training. When the network makes poor predictions early in training (inevitable with random initialization), the averaging prevents gradient explosion. Other geometric terms lack this property. Gradient-based normal estimation amplifies errors during early training. Edge detection produces sparse, high-magnitude gradients. Surface normal consistency introduces competing gradients across different spatial scales.

The Laplacian's 6-connected neighborhood also matches the problem scale. For $32^3$ voxel grids, each voxel represents roughly 9 centimeters. The immediate 6-connected neighbors capture local geometry without incorporating too much context. Larger neighborhoods would average over geometrically distinct regions, blurring important boundaries. We confirmed this empirically: 26-connected neighborhoods degraded performance by introducing diagonal connections that amplified discretization artifacts.

Implementation simplicity matters too. The discrete Laplacian requires just a 3D convolution with fixed weights. No complex gradient calculations, no numerical stability concerns, no hyperparameter tuning beyond the single weight $\lambda_c$. This simplicity accelerates both implementation and debugging.

\subsection{Limitations and Future Directions}

Our experiments focus on indoor scenes from NYU Depth V2. Outdoor environments with different geometric characteristics remain untested. The $32^3$ resolution, while practical, limits fine geometric detail. We only evaluate at 5\% depth sparsity; performance at 1\% or 10\% remains unknown. Perhaps most significantly, we compare primarily against vanilla VAE baselines rather than recent state-of-the-art methods like OccDepth~\cite{li2023occdepth} or MonoScene~\cite{zhang2022monoscene}. These comparisons would strengthen our claims but require substantial additional experimentation.

Preliminary tests at $64^3$ resolution show similar relative improvements (17.3\% over baseline), suggesting our approach scales. However, memory requirements grow cubically, limiting practical resolutions. Hierarchical approaches applying curvature regularization at multiple scales might address this limitation while maintaining computational efficiency.

The curvature regularization principle likely extends beyond VAEs to other 3D reconstruction architectures. Testing with diffusion models, GANs, or transformer-based approaches could reveal broader applicability. Adaptive weighting schemes that adjust $\lambda_c$ during training might further improve results, though this adds complexity we specifically aimed to avoid.

\subsection{Practical Implications}

Robotics navigation stands to benefit immediately. The 18.1\% reconstruction improvement translates directly to better obstacle detection and path planning. Smoother surface predictions reduce false positives from discretization noise, while preserved edges maintain critical boundary information. Training stability (coefficient of variation $< 0.5\%$) enables continual learning where robots adapt to new environments without extensive retraining.

Augmented reality applications require smooth surfaces for realistic virtual object placement and accurate physics simulation. Our curvature predictions identify stable surfaces (low curvature regions) for object placement and collision boundaries (high curvature regions) for physics. The 83 FPS inference speed supports interactive AR on mobile hardware.

Safety-critical autonomous systems need reproducible results with low variance. Our exceptional stability across random seeds (standard deviation 0.0021) addresses this requirement. The 22\% faster convergence (21 vs. 27 epochs) also reduces development iteration time when adapting to new sensors or environments.

\section{Conclusion}

Sparse depth reconstruction improves substantially through curvature regularization. Our discrete Laplacian approach achieves 18.1\% better accuracy than standard VAEs while maintaining exceptional stability and requiring only 15\% additional training time. These results challenge assumptions about geometric regularization in deep learning.

The broader lesson concerns mathematical simplicity. Geometric deep learning has trended toward complexity, assuming that combining multiple constraints improves performance. Our experiments demonstrate the opposite. A single well-designed regularization term, properly grounded in mathematical principles, outperforms more elaborate multi-term formulations. The discrete Laplacian succeeds through stable gradients, noise suppression, and scale-appropriate regularization.

We hope these findings encourage researchers to reconsider fundamental mathematical primitives before combining constraints. Simplicity, when properly justified, often exceeds complexity in both performance and training efficiency. Our code and models are available at \url{https://github.com/Maryousefi/GeoVAE-3D} to facilitate further exploration of these principles.

\section*{Acknowledgments}

We are grateful to Nathan Silberman and colleagues for making the NYU Depth V2 dataset freely available to the research community, enabling this work and countless other studies in 3D scene understanding.

\bibliographystyle{plain}
\bibliography{references}

\end{document}